\documentclass{article}

\usepackage{lingmacros, fullpage, setspace, amsmath}     

\usepackage{arxiv}

\usepackage[utf8]{inputenc} 
\usepackage[T1]{fontenc}    
\usepackage{hyperref}       
\usepackage{url}            
\usepackage{booktabs}       
\usepackage{amsfonts}       
\usepackage{nicefrac}       
\usepackage{microtype}      
\usepackage{lipsum}		
\usepackage{graphicx}
\usepackage{natbib}
\usepackage{doi}
\usepackage{hyperref}       

\newcommand{\bs}{\bigskip}

\newcommand{\noi}{\noindent}
\usepackage{xcolor}
\definecolor{color5}{HTML}{006795}
\hypersetup{
  colorlinks   = true, 
  urlcolor     = blue, 
  linkcolor    = color5, 
  citecolor   = color5 
}

\begin{document}
\setlength{\parindent}{15pt}

\title{Are language models rational? \\ The case of coherence norms and belief revision}

\author{ 
	Thomas Hofweber \\
	Department of Philosophy\\
	University of North Carolina at Chapel Hill\\
	\texttt{hofweber@unc.edu} \\
	\And
	Peter Hase \\
	Department of Computer Science\\
	University of North Carolina at Chapel Hill\\
	\texttt{peter@cs.unc.edu}
	 \And
	Elias Stengel-Eskin \\
	Department of Computer Science \\
	University of Texas at Austin \\
	 \texttt{esteng@cs.utexas.edu} \\
	 \And
	 Mohit Bansal \\
	Department of Computer Science \\
	University of North Carolina at Chapel Hill \\
	\texttt{mbansal@cs.unc.edu} \\
}

\maketitle

\begin{center}
{\bf Revised version of \today}
\end{center}

\bs
\bs

\setstretch{1.618}

\begin{abstract}

Do norms of rationality apply to machine learning models, in particular language models? In this paper we investigate this question by focusing on a special subset of rational norms: coherence norms. We consider both logical coherence norms as well as coherence norms tied to the strength of belief. To make sense of the latter, we introduce the Minimal Assent Connection (MAC) and propose a new account of credence, which captures the strength of belief in language models. This proposal uniformly assigns strength of belief simply on the basis of model-internal next token probabilities. We argue that rational norms tied to coherence do apply to some language models but not to others. This issue is significant since rationality is closely tied to predicting and explaining behavior, and thus it is connected to considerations about AI safety and alignment, as well as understanding model behavior more generally. 
\end{abstract}


\section{Introduction}

The question whether an AI is rational is of fundamental importance for understanding, explaining, and predicting its behavior. Rationality is a key aspect of what human beings generally attribute to each other when we try to explain and predict our behavior. But it is unclear whether anything like this is also applicable to machine learning models and AI systems more generally. In this article, we consider the question for a specific case. First, we will focus on a particular important class of AI systems --- language models --- and, second, on a particular important part of rationality: coherence norms and belief revision. Thus, our main question will be whether language models fall under this central class of rational norms. 

To get clearer on our main topic, it is important to first note that the term `rational' is used in different ways, and depending on how one understands it, our main question either becomes a trivial one or a more substantial one. On the one hand, `rational' is often used simply as a term of approval, where an AI is rational just in case it does what it is supposed to do. See \cite{russell:rational}. On such an understanding, intelligent systems are rational, as long as they work properly. But this conception of being rational does not consider how the system does what it is supposed to do and what internal states of the system contribute to its success. A different sense of `rational' focuses on these internal states instead, not just on the overall behavior of the system. This is how we often evaluate human beings as rational. Here, we do not simply evaluate their behavior, but we focus on their internal states and evaluate them. We consider their beliefs, reasoning, or actions as rational, and we thereby evaluate them as having arisen in the proper way from the proper internal, representational states. We can call the former, thinner notion of being rational the {\em deflationary} notion, and the latter the {\em representational} notion, since it concerns internal representational states. 

Exhibiting intelligent behavior and being rational are closely connected in the deflationary sense of rationality, but they can come apart when we compare intelligent behavior and being rational in the representational sense. Focusing on the representational notion, we get a substantial question about whether an AI is rational in this sense, even if it exhibits intelligent behavior. The question then comes down to whether its intelligent behavior is properly related to and explained by its internal representational states which bring it about. This is how our human intelligent behavior is commonly explained: by reference to our beliefs and desires, which represent the world and guide our actions. A (representationally) rational AI will thus be similar to our human intelligence in at least this way. But on the other hand, if the intelligence of an AI is not to be explained by it having representational states that are subject to considerations of rationality, but via some other, non-representational, mechanism, then the intelligence of this AI will be very different from our human intelligence. Such an AI might manipulate symbols that have no semantic meaning, or it might push matrices through a neural network without those matrices representing the world around it, or do something different altogether. In that case, it will be very different to explain, predict, and understand it. This in turn will affect our attempts at AI safety, alignment, explainability, and so on, and thus this issue is of significant interest. 

We will in the following focus on representational rationality, in particular for language models. Below we will focus on auto-regressive language models like Claude, ChatGPT, Gemini, and many others. It will not matter for our discussion just which of the more recent versions of these models we consider, since our discussion will be fairly abstract and general, targeting a general class of language models that they all belong to. We will investigate such models with an eye on a special kind of representational state that is subject to rational evaluation: beliefs. Other representational states, like intentions or desires, can also be subject to rational evaluation, as can be a combination of different states. An intention can be irrational given what one believes and what one desires. However, considering the rational evaluation of belief first, and leaving more complex cases of the rational evaluation of a mixed group of states for later, is a reasonable first step in understanding the place of rationality in machine learning models. Beliefs are the most paradigmatic and simplest kind of states subject to rational evaluation and thus a good candidate to consider first. Similar issues will arise for other representational states, and combinations of states. As we will see below, there are a number of different rational norms that can be seen as applying to belief alone, with the simplest ones being coherence norms, which are our main topic. 

The question of rationality can arise both for belief states over time and for belief states at a time. The former, {\em diachronic rationality}, concerns how belief states are updated over time, for example in light of new evidence, reasoning, or some other change the system undergoes. The latter, {\em synchronic rationality}, concerns the relationship between beliefs that are had at the same time. Coherence norms most directly relate to synchronic rationality. The simplest way for a system of beliefs to be incoherent, and thus to violate a coherence norm, is for it to contain contradictory beliefs at the same time: the belief that Paris is in France and at the same time also the belief that Paris is not in France. One's overall set of beliefs should cohere and not include any contradictions, or so rationality requires. It requires that one's beliefs collectively go with each other, and if they do not, to change them to reach such a coherent state. This requirement might not include a recipe for how to change them, only the demand that they need to be changed, since having this incoherent set of beliefs is not how things should be. This is a paradigmatic example of a coherence norm, and such norms are our focus below. 

In the literature on language models, there is much discussion about whether language models have the ability to reason, which is a cross-temporal rational ability. See \cite{huang2023reasoning}. Reasoning involves a sequence of steps that are properly rationally supported by earlier stages in the process. But the question whether language models engage in reasoning is hard to properly address without first settling whether language models are rational at all and whether norms of rationality even apply to them. The simplest case for making progress on that basic question is to focus first on synchronic coherence norms and on belief states governed by such norms. 

Thus to ask whether language models are subject to norms of (representational) rationality like coherence norms involves asking first whether they have the proper internal representational states like belief states that are in principle subject to such norms, and secondly whether the models with those states indeed are subject to such norms. To get clear on this issue, it is important to distinguish two ways in which a language model can fail to be rational. The first concerns whether a language model is under the norms of rationality in the first place. Here `rational' contrasts with being {\em arational}: not being subject to the norms of rationality. The second concerns whether the model lives up to the norms of rationality that apply to it. Here `rational' contrasts with being {\em irrational}: not living up the norms of rationality that apply. A toaster is not rational in the first sense, while a person with contradictory beliefs is not rational in the second sense. An arational toaster does not fail to live up to the standards of rationality, since those standards do not apply to it at all, while an irrational person does fail this way. We need to ask both of these questions of language models: do the norms of rationality apply to them at all, and if so, do the models live up to those norms. In this paper we will focus mostly on the first, more normative or philosophical question. In a different paper, \cite{hase2024fundamentalproblemsmodelediting}, we consider the empirical question, the results of which will also be discussed below.

Thus overall we need to make progress on the following questions: a) whether language models have the right kinds of internal states, like beliefs, that are candidates for being subject to the norms of rationality, b) whether coherence norms do indeed apply to those internal states, assuming language models have them, and c) whether the models live up to the norms that apply to them, assuming any apply. Progress on these issues should help us with the question whether language models are arational, irrational, or rational. We will consider these three issues in turn now, with an emphasis on the first two.

\section{Belief in language models\label{sec:belief}}

The easiest way to argue that language models have beliefs is via the connection of belief to knowledge. Language models seem to know a lot about the world. Although they might make things up and get things wrong, they nonetheless often reliably produce correct answers to many questions about the world. For example, many large language models can correctly answer the question of what the capital of France is and thus are naturally taken to know that the capital of France is Paris. But knowledge is just a special kind of belief: a belief that is true and especially well supported. The details of the connection between knowledge and belief have long been controversial in the philosophical literature. In particular, it is controversial what else is required besides a state being a belief state that has a true content for this state to be knowledge. Nonetheless, the general connection between knowledge and belief relied upon here is almost universally accepted: to know is to believe. Thus, since language models know about the world, they have beliefs about the world.\footnote{For a discussion of the relationship between belief and knowledge, see \cite{sep-knowledge-analysis}.} 

But this argument is too simple to be persuasive. We must distinguish the model knowing about the world, on the one hand, from us coming to know about the world by interacting with the model, on the other hand. The latter can be possible simply because the model has the relevant information stored somewhere in its internal states, without those internal states themselves being states of knowledge. To illustrate, we can come to know about the world by interacting with a dictionary, but it seems dubious that the dictionary itself knows about the world. It merely carries information about the world. Or to make the point more forcefully, I can come to know about the local wildlife by looking at the tracks in the snow, which carry information about recent wildlife crossings, but it is mistaken to think that the snow itself knows about the wildlife. 

To have beliefs about the world requires an internal state that is a state of believing, and simply carrying information alone is not enough for this. The question is whether language models have internal states which are states of belief. There are two main worries about this, one is very familiar and one less so. The first concerns whether the internal states of language models are representational states at all. Do any internal states of the model represent the world which created the text on which they are trained? Or does the model merely reliably predict text without any regard to the semantic properties of the text they produce? This issue is widely discussed for language models\footnote{See \cite{bender-koller:climbing}, \cite{piantadosi-hill}, and many others.} as well as for our human minds, where it is often phrased as the problem of intentionality or content.\footnote{See \cite{brentano1995psychology}, \cite{fodor1987psychosemantics}, and many others.} This problem concerns whether and how internal states of a system represent or are about external states of affairs. We do not and cannot hope to resolve this problem here. Instead, we will grant that the internal states of the models are representational and are about external things or states of affairs, and then focus on a different problem tied to models having beliefs about the world. This problem is not as widely discussed and arises even if we assume, as we now do, that models have representational states about the world. The worry is that none of these representational states are belief states since they do not meet the minimal requirements for a state being a belief state. This problem is the focus of this section.

Not all representational states are states of belief, even in minds that have beliefs. Beliefs are a distinct kind of representational state, and their distinct nature is revealed by focusing on the standards of correctness that govern beliefs. Different kinds of states come with different standards of correctness and what those standards of correctness are is revealing about the nature of the corresponding states. If a state does not live up to its standard of correctness, then there is something defective about it. Belief has a special standard of correctness: truth. A belief that is not true is thereby defective. Being true is what a belief is supposed to achieve, and if it does not, then that alone makes clear that it does not do what it is supposed to do. 

Not all representational states have this standard of correctness; belief is special in this regard. To illustrate this point, consider the act of imagining. Imagining is a representational state just like belief, but it does not have truth as its standard of correctness. If I imagine that I am the head chef at a French restaurant, then simply because this is false, there is nothing wrong with this imagining. But if I believe that I am the head chef at a French restaurant, then this belief is defective simply because it is false. There might be all kinds of benefits that come with having this false belief, such as increased confidence in the kitchen, but as a state of belief it is defective simply because it is false. This issue is often summarized in the slogan that truth is the norm of belief or that belief aims at the truth: it is in the nature of this internal state that truth is part of its standard of correctness.\footnote{The connection of belief to truth is widely discussed in the philosophical literature. See, for example, \cite{williams:deciding}, \cite{railton:truth},\cite{velleman:aim}.} 

In light of this it is arguable that no internal state of a language model is a belief. Language models are machine learning models, after all, and their internal states are thus the internal states of machine learning models. Such models have one main task: to learn and model the data on which they are trained. In the case of language models, that data is text. What the model aims to model is how text in a large dataset of text continues. In particular, the model assigns probabilities to the next token, basically the next word, given an initial sequence of tokens/words. The goal of the model is to capture what probabilities are supported by the data on which they are trained. To achieve this the model is gradually optimized over the course of the training process to assign probabilities to the distribution of token sequences that corresponds to the actual distribution of those token sequences in the training data. Since the target of this process of optimization is to assign probabilities as close as possible to the distribution of tokens in the training data, this is naturally taken to be what the model is supposed to do overall.  

Correspondingly, this is what the internal states of the model are supposed to help achieve. The internal states of the model are supposed to help with this assignment of probabilities to the next token. And those internal states can be better or worse at doing that. Internal states that contribute to the wrong probabilities are defective, and those that contribute to the right probabilities are doing what they are supposed to do. But on this way of looking at it, the correctness conditions for internal states of language models are simply determined by how well they do in contributing to what the model is supposed to do: model the training data. In our present case, it is supposed to capture the probabilities of the next token, and that's all. Thus the standards of correctness of the internal states of language models as machine learning models are tied to the data on which they are trained, but not directly to the world in which this data was generated. In particular, the model does not have internal states that have truth as its standard of correctness. And thus none of its internal states are beliefs. 

This argument against language models having beliefs is different from the argument that their internal states are not representational. Even if the internal states were representational, the question remains whether any of these representational states are beliefs. And that question is closely tied to the question whether any of the internal states of the model have the standards of correctness that are definitive of a belief state: truth. Even if the internal states of the model represent the world, their standard of correctness seems to be exhausted by how well these states contribute to the success of the model as a machine learning model. And as a machine learning model, the target is to capture the data. Representational states might help with capturing the data, but the correctness of these states would be tied to them capturing the data, not the truth of their representations. As it stands, no such state can be a belief. 

The argument given applies even if the machine learning model builds a world model to learn the data. A language model might build a world model, that is, a representation of the world and the agents that produce the text it is supposed to learn, and it might be the most efficient way of learning the data. Such a world model would be representational, and it might well be true. But its standard of correctness is still determined by being instrumentally useful for learning the data. The data might be learned most efficiently by a world model that is not an actual representation of the world, say because the training data is best learned by building a world model which models the world in an inaccurate, but simpler, way. That world model would be false, but it would do what it is supposed to do: help learn the data. A true world model might in general help learn the data better, but the internal representational states of that world model are not flawed simply because they are false. They would be flawed if they contribute badly to learning the data. Thus even representational states that make up a world model are not beliefs. Their instrumental goal is still learning the data, not the truth. Being a truthful representation might, and likely will, help learn the data. But the standard of correctness of these internal states is not truth itself, only learning the data. And thus even world model representations by themselves are not beliefs. 

This argument so far has force, but it is not the end of the issue. Even if this argument against beliefs in language models is perfectly correct, it does not rule out beliefs in all language models. Terminology in this area is not always used uniformly, but we will use the following terminology to make an important distinction: To assess the above argument, we need to distinguish {\em pretrained} language models from both {\em fine-tuned} language models and from {\em grounded} language models. A pretrained model is simply trained on text and nothing else. It is the kind of language model relied upon in the above argument. All it does is assign probabilities to the next token on the basis of learning from the distribution of tokens in the training data. However, pretrained language models are often only the basis of other language models, not the end of language models. Language models we often interact with, i.e. autoregressive models like ChatGPT, Claude, or Gemini, are not simply pretrained models. They are further fine-tuned beyond their pretraining. They can be fine-tuned for a question-answer format, or on especially significant texts for a particular application. But this fine-tuning can also happen in other ways than training on more text. One widely used technique is to train the model further by employing reinforcement learning from human feedback (RLHF). Here human beings are tasked to rate outputs of the language model, which is used to train a reward model, which in turn is used to adjust the weights of the original pretrained language model to achieve a higher reward. The key now is that this human feedback can reward the model's output along different dimensions than simply getting the probabilities of the next token in the training data correct. The probabilities for the next token change from those initially learned from the training data to those that lead to a higher reward. And those new dimensions leading to a reward can be as diverse as being helpful, being concise, being elegant, and most importantly for us here: being truthful. Human feedback thus can reward truthfulness in the responses. And since this is used in fine-tuning the model, the internal states of the model do therefore not solely aim to contribute to the proper assignment of probabilities of next token based on the training data in pretraining, but they also aim to achieve the reward tied to fine-tuning, which includes truthfulness. Through this fine-tuning the internal representational states of the model can acquire a new goal --- being truthful --- and with it they can become beliefs.

Thus a pretrained model might not have beliefs even if its internal states are representational, since none of its internal states have truth as part of its standard of correctness, while a properly fine-tuned model can have beliefs, since during the process of fine-tuning the internal states can be selected for their contribution to truthfulness, and thus truth becomes part of their standard of correctness. Fine-tuning a pretrained model with RLHF can achieve this, but it might also not be the only way to achieve it. The two other natural ways to do this is not to focus on the end of the process, i.e. fine-tuning a pretrained model, but on the beginning or the middle instead. The former is related to curating  the data on which the model is pretrained, and the latter concerns the nature of the model itself. We should consider them in turn, starting with the latter one. 

A model can be connected to the truth in other ways than fine-tuning for truthfulness via RLHF, and consequently its internal states can acquire the goal of truthfulness in other ways as well. The most obvious example is a model that is not a pure language model, but a multimodal model that is ``grounded''. Such a model can be connected to the world via perception. It might be connected to a camera that feeds it images as input. And these images in turn can affect its output and its internal states in a way that makes it responsive to the truth, just as our perceptual states affect our internal states. In particular, the internal representational states of the model can then be responsive to perceptual input just like our belief states are responsive to what we see and hear. Assuming that perception generally is veridical, or at least has the function to be veridical, the internal representational states of the model can then be sensitive to the truth if they are properly connected to the perceptual system. They would then not merely represent the world, but become beliefs. 

This sense of ``grounding'' the model in the world via perception is a different sense of ``grounding'' the internal states in the world for them to be representational states at all. The latter sense concerns whether the internal states properly represent the world at all. The former sense concerns making the internal representational states responsive to perceptual input. We simply grant here that the internal states of the models can be representational. This issue clearly deserves further discussion\footnote{For more on `grounding' in the sense of determining representational content and the significance of RLHF for it, see \cite{mollo2025vectorgroundingproblem}.} but even if we grant that the model has internal representational states that does not resolve the issue presently at hand, namely whether these states are aiming at the truth and thus might be beliefs. Proper perceptual grounding can assure this. 

The aim of having truthful representations can also be built in from the beginning. For a model that is simply pretrained on some text, we argued that its internal states do not thereby acquire truth as their standard of correctness. But if the model is trained on text that is especially selected for being truthful, then one could argue that the internal representational states have the function of being truthful and thus can be beliefs. Here it would not be enough that the training data consists largely of true text. This might be so by accident, with no effect on the overall function of the internal states. But if the training data consists of true text by design, if being trained on true text is part of the purpose of the model, then it is arguable that part of the purpose of the internal states is to be truthful, and thus that they can be beliefs. The issue is one about the standards of correctness of these internal states, which concerns what they are supposed to do or what their function is, not merely what they happen to do in fortunate circumstances. Thus, the difference between being trained on text that is selected for being truthful and being trained on text that happens to be true matters. It is the function of the states and what they are supposed to do that makes the difference. 

What precisely is required for internal states to have truth as the standard of correctness is unclear and more needs to be said about it elsewhere. For example, it could be argued that in reinforcement learning from human feedback we do not get as close of a connection to truth as would be required, since the real connection is to what is believed to be true by the human giving the feedback. Similarly, multimodal models grounded in the world with a camera might not be connected to the truth directly, only to what is presented as being the case by the sensory modality. These issues also lead to well-known philosophical problems about how our human minds are connected to the world and whether our internal states are properly connected to the truth. After all, does perception connect us to the world directly and to the truth, or only to how the world appears to us? These are all issues worth pursuing, but for our present purposes we would like to conclude this section by stating that pretrained language models as such do not generally have beliefs, while fine-tuned and grounded models, as well as models trained on data curated for truthfulness, might well have beliefs. And once we accept beliefs, we are ready to approach our main question whether such states are under the norms of rationality, in particular coherence norms.

\section{Coherence norms for language models}

Belief in humans is subject to the norms of rationality. We are not always rational, but we ought to be, which is to say: we are under the norms of rationality, even if we don't always live up to those requirements. Rationality itself is complex, including different branches or domains of rationality, each with its own set of norms. It is common to distinguish epistemic from practical rationality, with the former concerning what one ought to believe and the latter concerning what one ought to do. Our focus here is on epistemic rationality, but practical rationality might also be significant for language models, since such norms govern what one ought to do generally, including what speech acts one ought to perform in light of one's beliefs and intentions. This form of rationality is widely discussed in speech act theory,\footnote{See \cite{grice:studies} and \cite{austin1962how} for classic texts, and \cite{lorini:design} for an application to language models.} but not our topic here. In the following, we will not concentrate on what a model should do, assuming it has agency at all, nor on what it should say, assuming it makes assertions at all. Instead, we will focus on epistemic rationality: what the model should believe. 

The norms of epistemic rationality can themselves be analyzed as a series of distinct norms. One core part of these norms are coherence norms, which concern how the internal states of a system ought to relate to each other. Other norms of rationality concern the proper response to reasons or evidence. How the different parts of rationality relate to each other is itself controversial, but we will simply focus on coherence norms in the following.\footnote{See \cite{kolodny:rational}, \cite{kiesewetter:norm}, \cite{worsnip:fitting} and many others for the debate about whether coherence norms are derivative on norms tied to responding to reasons or whether they are an independent group of norms of rationality.} 

Coherence norms come in different kinds, most notably as ones concerning full belief and ones concerning degrees of, or strength of belief. Most notably among the former are norms concerning logical coherence and among the latter are norms about probabilistic coherence. Logical coherence norms concern the relation of our beliefs to each other with respect to logical implication and consistency. To give a simple example, a logical coherence norm might demand that one ought not to believe a particular proposition $p$ as well as its negation $\neg p$. Probabilistic coherence norms in general also come in different kinds. One kind concerns belief about probabilities itself. For example, a norm might demand that one ought not to believe that the probability that it is going to rain is 90\%, and also believe that the probability that it is not going to rain is 60\%. But besides beliefs explicitly about probabilities, there are also coherence norms tied to subjective probabilities or credences. Credences correspond to how strongly one believes a particular proposition. So, to believe that Paris is in France with credence 99\% is to believe it very strongly, but it is not itself a belief about probabilities. Therefore to believe that Paris is in France with 99\% credence is not the same as to believe that the chance that Paris is in France is 99\%. Credences should be coherent, and the view that they should satisfy the axioms of probability theory is generally called {\em probabilism}. For example, probabilism holds that if your credence that $p$ is $x$, then your credence that $\neg p$ ought to be $(1-x)$, for $x \in [0,1]$. We will not discuss coherence norms tied to beliefs explicitly about probabilities in this paper, but focus on logical coherence and the coherence of credences instead. 

To hold that language models ought to satisfy these coherence norms is problematic in at least two ways. 

First, it is unclear why logical coherence should be required of language models. After all, they are machine learning models and thus they are supposed to learn the training data. But the training data for language models is generally incoherent. Language models are trained on text from the internet and other sources, which is not a coherent unified data set of text. Since the machine learning model is trained on incoherent text as its training data, why should we demand of it that it be coherent? In fact, coherence seems to go against what it is supposed to learn from the training data. Incoherent training data should lead to a model with incoherent beliefs if the model properly captures the training data. Thus to require the model to be coherent seems to be unjustified, and therefore the rational norm of coherence does not seem to apply to it. Unless the training data is coherent, the model should not be required to be coherent either. This contrasts with why coherence norms apply to human beings. We learn from a coherent world: a world without any contradictions. Thus our internal states should be coherent, to properly reflect what our minds are trained on. The world is mediated to us by evidence. But even if our evidence is conflicting, pointing partly in one direction and partly in another one, what the evidence points to --- the world --- is coherent. Our primary goal is not to learn the evidence, but to learn about the world from the evidence. That is why we need to resolve conflicting evidence. Thus the norm of logical coherence applies to us, since we aim to represent a coherent world. But no analogous case carries over directly to language models, whose primary goal is to learn the training data, not about the world. 

Second, it is unclear how probabilistic coherence of credences can be required of language models since it is unclear how to make sense of credences for language model beliefs in the first place. Credences are an assignment of probabilities to propositions that correspond to the strength of a belief. And although language models are heavily involved with probabilities, those probabilities are not probabilities of propositions and they do not correspond to the strength of a belief. For example, a model will assign probabilities to the next token given a sequence of tokens, but this assignment is an assignment to tokens, not propositions. And it is an assignment of how likely the next token is to appear in a text, not an assignment that corresponds to the strength of a belief. Thus even if language models have beliefs, it is unclear how to understand how strongly they believe a proposition. To hold language models to the standards of probabilism, and with it to claim that they ought to satisfy the corresponding norms of coherence for credences, only makes sense if credences themselves make sense for language models.

Both of these problems have a solution, and we should look at them in turn. 

\subsection{Logical coherence norms}

The solution to the first problem is similar to the argument given above in Section \ref{sec:belief} for the coherence of beliefs in language models. It is true in general that a pretrained language model should not be seen as being under the norm of logical coherence. Such a model is simply aiming to model the training data, and that data is not logically coherent. To require coherence of a pretrained language model that is trained on incoherent data is to require something of it that goes against what it fundamentally aims to do: learn the training data. But things are again different for fine-tuned models, in particular ones that are fine-tuned for truthfulness. Properly fine-tuned models can acquire the function to produce truthful text, and if we assume that such models have internal representational states like beliefs and that those states are properly causally connected to the text they produce, then the aim of being truthful can support that logical coherence norms apply to the internal states of such models. Internal belief states that are incoherent will be contrary to the norm of truth which is part of the standard of correctness for belief itself. Thus inconsistent beliefs are in conflict with the standard of correctness of belief, and thus the coherence norm of consistency for beliefs applies. A pretrained model does not need to be under such norms, but a model fine-tuned for truthfulness is.

The norm of logical coherence is thus derivative on the norm of truthfulness. Any system that aims at truth must also aim at coherence, derivatively, since it cannot achieve its primary goal of truth without also being coherent. The truth norm is thus stronger than the coherence norm. Truth concerns the relationship between beliefs and the world. Coherence concerns the relationship between beliefs alone. The truth norm gives rise to the coherence norm since the world is coherent, and thus truth cannot be achieved without coherence. But coherence alone is not sufficient to characterize belief and what is distinctive of what kind of a state belief is. Beliefs that are perfectly coherent with each other but false are defective belief states: they are not how things should be and what is aimed for with these states being states of belief. A false belief is not how it should be, even if it is coherent with one's other beliefs. The norm of truth is thus primary and the norm of logical coherence is derivative on it. Nonehteless, both norms apply, to use and to properly fine-tuned language models.

\subsection{Defining credence for language models}

Logical coherence norms make sense for certain language models. But do coherence norms tied to credences also make sense? Those coherence norms are tied to the strength of belief, and how the strengths of different beliefs are supposed to relate to each other. This issue is simply not addressed by focusing on logical coherence, which concerns full belief and whether one's full beliefs are contradictory.  To make further progress on needs to overcome one main obstacle: how to make sense of credences for language models in the first place. Credences correspond to how strongly something is believed. But how is strength of belief to be understood for a language model? To make sense of it we need to assign probabilities to propositions, which are the contents of what is believed. This assignment of (subjective) probabilities in turn must correspond to the strength of belief: the closer the number is to 1, the stronger the corresponding proposition is believed, which in turn means that it is harder to give that belief up in light of new evidence, or in light of a conflict with other beliefs that are assigned lower numbers by this credence function. To make sense of such an assignment of probabilities to propositions for language models is the key first step to making progress on the question whether coherence norms tied to credences apply to language models. 

Language models are steeped in probabilities, but those probabilities concern the next token, not the strength of belief. Tokens generally are close to a word and thus sub-propositional. But credences apply to propositions as a whole, which are close to a complete sentence. The question is how can we make use of probabilities like $$P(t_n | t_1, ....., t_{n-1}) = x$$ which are assigned by the language model to the n-th token given that the previous tokens are $t_1, ..., t_{n-1}$, and turn that into an account of what the model's credence in proposition $p$ is. 

We propose to solve this problem via the {\em Minimal Assent Connection} (MAC). The basic idea of MAC is simply to consider the probability of the model predicting the next token to be `yes' when asked whether it is the case that $p$. This way next token probabilities can be directly tied to probabilities assigned to propositions. To put a label on it, we can say that the model {\em minimally assents} to $p$ just in case the probability it assigns to `yes' being the next token after `Is it the case that $p$?' is sufficiently high. This basic idea is promising, but to in order to turn it into a credence function that captures strength of belief more needs to be said and we will do this momentarily. But on the positive side, we can already note that this approach is completely general, since for any proposition $p$ we can consider the probabilities that the model assigns to the next token after the question whether it is the case that $p$. This probability comes directly from the language model. Next token probabilities are given in the output layer of the model and are thus easily accessible and require no special knowledge of the more hidden, internal layers of the model.  This is just the kind of assignment of probability we want to exploit in assigning credences to the model. This assignment applies to any proposition $p$ uniformly. Furthermore, the probability of a `yes' answer to the question whether $p$ is naturally associated with the strength of believing $p$. After all, it is natural to hold that the probability assigned to `yes' after the question whether $p$ is closely correlated with the strength of believing $p$: the stronger the confidence of the model in $p$, the higher the probability for a `yes' answer, all things being equal.  But still, it can't quite be a full answer and we can't make this connection as outlined. That is, we can't fully endorse that we can define $cr(p)$, the models credence in $p$, as follows:

\begin{equation}
cr(p) = P(\text{Yes } \mid \text{ Is it the case that } p?)
\label{cr}
\end{equation}

There are essentially two flaws with this proposal. First, the model might assent to $p$ in ways other than with `yes'. It might also assent via other forms of affirmation, say `yeah' or `indeed'. Using just (\ref{cr}) thus gives a credence that is too low. But we can simply fix this in one of two ways. First, we could force the model to answer either `yes' or `no', either by adding this requirement to the question prompt, or by constraining the model in other ways to do so (e.g. by constraining and re-normalizing the output distribution). However, adding such further constraints slightly changes the model or the text for which it predicts the next token. Making such a demand for either a `yes' or a `no' answer thus requires changes that make the method not completely general any more. It would not apply to any model any longer, only to ones properly modified to meet this constraint. But there is also a second way to solve this problem. We can leave the model and input alone, and instead simply sum over all affirmations, not just `yes', but also `sure', `yeah', and so on, including multi-token sequences like `I couldn't agree more.' All such tokens or sequences of tokens we can label an {\em assent sequence}. If we call $AS$ the set of all such assent sequences, then we can better define the credence for $p$ by the model as the sum of all conditional probabilities of assenting. So, we can define the {\em assent probability} as:

\begin{equation}
as(p) = \sum_{s \in AS} P(\text{s } \mid \text{ Is it the case that } p?)
\label{cr2}
\end{equation}

We can then, as a second attempt, identify the credence of the model in $p$ with $as(p)$. Setting the credence of $p$ to $as(p)$ is an improvement over (\ref{cr}). But it isn't quite right either, since it again generally underestimates the credence and sets it too low. The reason for this is simply that the model might answer the question whether it is the case that $p$ in some way completely unrelated to its belief in it. It might answer `As an AI language model I cannot address issues of this nature' or some other way that avoids answering this question. Some of the probabilities for how to continue the sequence will be allocated to such answers, and thus the part of the probability distribution that is devoted to assent sequences does not properly represent the strength of belief of the model. 

But we can fix the approach to overcome this difficulty. Strength of belief can be associated with how likely the model is to minimally assent when compared to how likely it is to minimally dissent, leaving aside other ways to respond to the question besides assenting or dissenting. So, we can simply consider the ratio of assenting to either assenting or dissenting. To make this idea work, we also need to consider the set of all dissent sequences $DS$, which are the ways the model might minimally dissent: `no', `never', etc.. 

\begin{equation}
ds(p) = \sum_{s \in DS} P(\text{s } \mid \text{ Is it the case that } p?)
\label{cr3}
\end{equation}

So understood, we should not expect that $as(p)$ and $ds(p)$ sum to 1, since the model might respond in other ways than assenting or dissenting. But we can simply renormalize by considering only the ratio of probabilities given to affirmations to those that are given to either affirmations or dissent, leaving out all other options. And this ratio is the proper way to assess the strength of belief of a model in a proposition $p$: how likely it is to assent when we consider only assent or dissent, leaving out all other options. This then is the proposal for how to assign credences to a language model in a way that is derivative only on the model's own assignment of probabilities to the next token:

\begin{equation}
cr(p) = \dfrac{as(p)}{as(p) + ds(p)}
\label{cr4}
\end{equation}

This association properly corresponds to the strength of belief in a proposition by the model, assuming we can talk about belief in the model at all. It naturally corresponds to the idea that the more likely the model thinks that the next token is `yes' following the question whether $p$, the more strongly it believes that $p$.

Articulations of what the assent and dissent sequences are will approximate the target credences of the model, with the proper articulation reaching its target. But for practical purposes it might well be enough to work with what we can call the {\em Yes-No approximation}: which considers only `yes' as an assent sequence, and only `no' as a dissent sequence. Thus the Yes-No approximation to a model's credence assigns to a proposition $p$ the credence which is the ratio of the model's next token probability assigned to `yes' after the prompt `Is it the case that $p$?' divided by the sum of the probabilities for `yes' as well as for `no' (conditional on the same prompt). This will approximate the true credence, but it is an empirical question how closely for a given model. 

We have not given a detailed list of assent and dissent sequences, and doing so requires one to make some choices that can be subject to debate. One concern is tied to context sensitivity. It could be that what counts as assent or dissent for a particular issue $p$ depends itself on $p$, and thus it is not clear whether assent and dissent sequences can simply be listed for questions concerning any $p$. Although this is in principle possible, it does not seem to be a major concern. First, it is unclear what plausible cases there are for this, and second it merely requires a reformulation of the proposal by relativizing assent and dissent sequences to a given $p$, and then consider the assent/dissent ratios concerning $p$ with each of those relativized to $p$ itself. 

A more difficult case is presented by examples of sequences that concern the epistemic state of the model itself, as `It seems to me to be unlikely,' `I am quite sure that $p$,' or `it is doubtful.' Such sequences contain `epistemic markers' or `epistemic strengtheners or weakeners' that report on the epistemic state of the model. Strong epistemic markers indicate that the model is in a strong epistemic position to make an assessment, as with `I am sure that...,' or `I am quite certain that..'. Weak epistemic markers indicate a weak epistemic state, as in `I am not sure, but it seems that...' or `I doubt that...'. Should sequences with such epistemic markers, either strong or weak, count as assent or dissent sequences for our purposes here? Should next token probabilities allocated to such sequences count towards assent or dissent probabilities?

On the one hand, it is clear that `I am sure that $p$' is a form of assenting to $p$, and that prima facie speaks in favor of including it as an assent sequence. But doing so would have a negative impact in a related area. When a model epistemically marks its response, then this marking should correspond to its confidence in the answer, and thus it should correspond to the credence it has in this being the right answer. So, when a model responds `I am certain that $p$' then this is the proper response only if the credence in $p$ itself is very high. One should only say that one is certain that $p$ if one has high credence in $p$. In fact, this connection can be seen as a further coherence norm for rational thinkers. If the strength of my epistemic marking diverges from my credence, then I am arguably violating a coherence norm. If, however, we include epistemic markers in the list of assent and dissent sequences themselves, then the credence of a model in $p$ will in part be determined by its epistemic markings of $p$. But these issues should at first be kept separate. It is one thing how strongly the model believes that $p$, and another whether a particular epistemic marking is appropriate for that model. It is thus overall best to exclude such epistemically marked sentence from the list of assent and dissent sequences. Maybe it is best to think of `I am quite sure that $p$' as containing two parts, made more explicit by its reformation as `$p$ and I am quite sure of it'. The first conjunct is a proper assent to $p$, while the second one is not, and only a comment on the model's own epistemic state. We should only count the first conjunct as an assent sequence, but not the combination of both as a further one. 

Whether or not present-day models' credences match their epistemic markings is an empirical question. If they do they are epistemically better functioning than if they don't. To determine whether there indeed is this correlation is subject to ongoing work. See also \cite{zhou-relying} for more on models reporting on their own epistemic state.

Furthermore, we would like to note that $$as(\neg p) = ds(p)$$ which is to say `the strength of assenting to $\neg p$ being equal to the strength of dissenting to $p$' is not a requirement on credence itself, but at best a norm of rationality tied to credences. Credence itself does not come with a requirement that assenting a negated proposition is properly correlated with dissenting the un-negated proposition. That there should be such a connection is reasonable to demand for any believer who is rational. But it should not be built into an account of what credence itself is for a believer, and it is not part of our definition of credence for language models. That particular models might not meet this requirement does thus not speak against the present account of credence for language models itself, but it would be a strike against the rationality of these models, all things considered. 

Finally, we would like to make clear that the assignments of credences to beliefs for models as defined above are for revealed belief, since they are based on the model's response to questions. This concerns belief as it is revealed by the model in its verbal behavior. Focusing on revealed belief does not mean that we take belief in general to be understood behavioristically.  Behaviorist theories of belief and other mental states see those states as derivative on one's behavior or dispositions to behave, rather than internal states that cause this behavior.\footnote{Classic behaviorist accounts of belief include \cite{skinner1957verbal} and \cite{ryle1949concept}.} The notion of revealed belief, that is, belief as it is revealed in one's (verbal) behavior, makes sense whether or not belief in general is understood behavioristically. This approach is limited in that it does not work for models that have the ability to lie and thus answer questions contrary to what they believe. It is not clear how one can make sense of lying in language models, since it is not clear how to make sense of belief in language models besides revealed belief. To be clear, lying here is understood as not just saying something that is false but saying something that is contrary to what one believes. This might be a temporary limitation to understanding belief in language models. If so, then our account of credences for revealed belief can be understood as an account of evidence we get from verbal behavior for the credences of beliefs more generally. Verbal behavior provides defeasible evidence for what one believes, and our account would provide evidence for how strongly a model believes something. But until an account of model belief has been developed that goes beyond revealed belief, we take our account of credences to be an account of the strength of belief in language models.

\subsection{The argument for probabilism}

Credences thus make sense for language models in general. As is common terminology, we can say that the {\em credence function} for a model is the function that assigns a proposition the corresponding credence that the model has towards that proposition. With the above account of credences for language models, a model's credence function is also well-defined. The question remains whether the norms of rationality tied to credences apply to language models, and with it the question remains what requirements there are on the credence function of the model. What kind of function should a model's credence function be?  

When focusing on synchronic coherence norms, this question quickly turns into the question whether the credence function should satisfying the axioms of probability theory and thus be a probability function. In the philosophical literature the view that a credence function should be a probability function is called {\em probabilism}.\footnote{See \cite{joyce:prob}, \cite{teitelbaum:bayes}.} Assuming credences for language models as spelled out above, does probabilism apply to them as well? Are language models under the coherence norm that their credence function satisfies the axioms of probability theory? For example, are they under the norm that $cr(p) = 1 - cr(\neg p)$?

To answer this question we first need to again distinguish pretrained language models from fine-tuned ones, in particular those fine-tuned for truthfulness. For a pretrained model we should not in general expect the model's credences to obey the probability calculus, nor that they are under any norm to do so. Since the model will generally be trained on incoherent text, we can not expect that $cr(p)$, the credence assigned to $p$, is exactly $1-cr(\neg p)$. If the training data is sufficiently incoherent, then it can well be that the credence for each of them is over 50\%. Since the model so far is merely a pure language model whose internal states are correct insofar they contribute to the training task and learning the data, we should not expect there to be a norm on these probability assignments that correspond to probabilism. 

However, things are again different for the model that is fine-tuned on truthfulness. A model that aims at true beliefs is subject to the well-known accuracy arguments for probabilism. See \cite{joyce:prob}, \cite{teitelbaum:bayes}. The main idea of these arguments is as follows: if the goal is truth, then credences should be as close to the truth as possible: one's credence in a true proposition should be close to 1, and one's credence in a false proposition should be close to 0. However close one comes to this ideal can be seen as the accuracy of one's credences. A standard measure of accuracy in this sense is the Brier score, basically the Euclidean distance between one's credences in some propositions and their actual truth values (taken as 1 for true, and 0 for false) seen as points in a Euclidean space of the dimension that corresponds to the number of propositions, but without taking a square root. So, the Brier score for a credence function $cr$ from a finite set of propositions $p_1, ... p_n$ to [0,1] is defined as 

\begin{equation}
BR(cr) = \big(tv(p_1) - cr(p_1)\big)^2 + ..... + \big(tv(p_n) - cr(p_n)\big)^2
\label{br}
\end{equation}

\noi whereby $tv(p) = 1$ if p is true and 0 if it is false.\footnote{The Brier score is sometimes also given as the mean squared error, i.e.~dividing the above by the dimension $n$. This difference does not matter for us here.} The thought is that a credence function is better if its Brier score is lower, since one aims at the truth, and a lower Brier score of one's credence function means that one's credences are closer to how they should be. In \cite{joyce:prob}, James Joyce showed that if one's credence function does not meet the axioms of probability theory, then there is a different credence function that has a lower Brier score no matter what the world is like and which meets the axioms of probability theory. However, if one's credence function does meet the axioms of probability theory, then this is not always the case. This argument is taken to support probabilism: A credence function that isn't a probability function can always be dominated by one that is. Thus such a credence function isn't what it should be, given that one aims at the truth. And thus probabilism is a rational norm that applies to one's credences: one's credence function ought to be a probability function.\footnote{For further discussion of Joyce's argument, see \cite{teitelbaum:bayes} volume 2, pages 338-71.} This argument applies to the credence functions of language models just as it applies to the credence functions of human beings. As long as we aim at the truth and at accuracy, i.e.~lower Brier scores, our credence functions ought to be probability functions, or so the norms of rationality demand. 

This then answers the question whether synchronic logical and probabilistic coherence norms apply to language models, or at least whether they apply to them just as much as they apply to us. We argued that such norms do not always apply to pretrained language models that simply model the distribution of words or tokens in a dataset of text. What is missing is the aim of being truthful. But this aim can be introduced in several ways, and with such an aim in place, the rational norms do apply as a consequence. In particular, fine-tuned language models which are fine-tuned for truthfulness are under such coherence norms. In that case, the norms of rationality apply to these models just as they apply to us. The next question to consider is whether this also holds for diachronic norms of rationality.

\section{Belief revision norms for language models}

The logical and probabilistic coherence norms we considered so far were synchronic norms: they apply to a system with beliefs at a time. Belief revision, on the other hand, is essentially diachronic and concerns how beliefs should be changed over time. There are different approaches to characterizing the norms that govern such rational change of belief. One is focused on full belief and a logical system of update rules that aim to capture such change.\footnote{See \cite{alchourron1985logic} for a paradigmatic example.} However, logic-based approaches are not naturally integrated with language models, as they do not operate primarily through symbol manipulation. It is much more natural when considering belief revision in language models to focus on norms that operate on credences, since, as we have seen above, those can be directly associated with belief in language models. This is what we did in the last section with synchronic coherence norms like probabilism. The most natural extension of probabilism to a diachronic norm is Bayesianism, which we here take to be the view that credences should be updated in light of new evidence in accordance with Bayes' Theorem. Rationality requires, according to a Bayesian, that one conditionalize on new evidence and adjust one's credences accordingly over time. 

This is a natural rational norm for belief revision as it applies to us humans and similar creatures. But it is not clear how it applies to machine learning models like language models. First and foremost, it is unclear how to make sense of the notion of {\em evidence} for a language model. For us human beings, evidence is naturally tied to perception by observing the world or to learning from others directly in communication. But language models by themselves do not have perceptual abilities. Even multimodal models do not have perception as such. A model trained on text as well as images from comic books, say, would not thereby have a perceptual ability, since the images are not properly connected to the world. Such a model is multimodal, but that by itself does not make it perceptual. Perception-like abilities could in principle be integrated with language models, say via feeding it images from a camera that depicts the world, but without a clearer understanding of how that will go, it is hard to make sense of evidence for language models and with it is hard to make sense of Bayesian updating and Bayesian approaches to belief revision. Understood as a diachronic norm of rationality, it does not seem to apply to language models as they are at present. 

Language models nonetheless can change their beliefs: they can be fine-tuned to adopt a particular belief or whole set of beliefs, and they can be directly edited to change a particular belief.\footnote{\cite{zhu2020modifying, de2021editing, dai2021knowledge, hase2021language}} But such change of belief of the model is not a rational process for the model itself. It is simply an external modification of the model, and the question whether that change in belief was rational does not directly apply from the point of view of the model. It is a change that happens to it from an external source, not a direct response to rational forces like evidence. But the question remains how the model should update its beliefs and credences in other propositions given that a particular belief has been changed. If a belief changes, then this will likely lead to incoherence and inconsistency unless other beliefs change as well. To preserve coherence, any change in belief cannot just be local and isolated, it must also affect many other beliefs that are connected to that belief. Such connections can come in the form of logical relationships or probabilistic ones, for example, conditional credences. 

Thus, even when a belief change happens in a non-rational way, and not as the result of responding to evidence, there are other norms of rationality that require further subsequent belief changes. Those norms of rationality can require that, given a particular new belief, various other beliefs need to be adjusted in light of how they are connected to the new belief. If the model now believes that the butler did it, then it should stop believing that the maid did it. And if it increases its credence that the cook did it, then it should decrease its credence that one of the other people did it.

This highlights that synchronic coherence norms generate rational requirements tied to belief change, even if that belief change was not itself done rationally. Even if the model is externally modified to hold a certain new belief, by fine-tuning or direct editing, the synchronic coherence norms require further changes in belief and credence to restore coherence or achieve it outright. Synchronic coherence norms can thus support conditional diachronic coherence norms, even if the model does not have the ability to respond to evidence. Conditional on having a given new belief, synchronic coherence norms require various revisions to its other old beliefs as well. 

What then do these norms demand for how the model's beliefs are to be modified? In general, there is no unique answer to this question. The easiest way to achieve coherence again after a belief change, assuming that coherence was there in the first place, is simply to undo the belief change and revert back to one's original state. But suppose that we keep the externally induced belief change and we will not revert back. Given that we fix the new belief and its credence, what do synchronic coherence norms require of the model to change as a consequence of this new fixed belief? What else needs to change, and how, to live up to the synchronic coherence norms? 

The most natural way to achieve coherence while keeping the new belief B fixed is to also keep conditional credences fixed as much as possible and adjust one's other beliefs in light of these conditional credences and the new fixed belief B. How this change should happen in a broadly Bayesian framework will depend on what credences one has in the new belief B. If it is considered fixed and certain, then B functions in effect like Bayesian evidence and one simply needs to adjust one's credences in any other belief A in accordance to one's prior conditional credences in A given B. That is to say:

\begin{equation}
    P_{new}(A) = P_{old}(A | B)
\end{equation}

If we, on the other hand, treat B not as certain and give it a regular credence below 1, then the most natural way to achieve coherence is to weigh the conditional credences relating A and B by the new credence in B. This corresponds to Jeffrey conditionalizing:

\begin{equation}
P_{new}(A) = P_{old}(A|B)P_{new}(B) + P_{old}(A|\neg B)P_{new}(\neg B)
\label{jc}
\end{equation}

These are widely discussed methods for updating one's credences \cite{teitelbaum:bayes}, and both are, of course, not without their critics. We do not have to settle the issue here as to which method is best. What matters for us here is not which method precisely is the correct one, but rather that there are derivative broadly Bayesian diachronic norms connected to belief revision that apply to language models and which are derivative on the synchronic norms defended above. Even though language models of today cannot yet be said to respond to evidence, nonetheless belief revision norms make sense for them. Externally induced belief changes via fine-tuning or other model editing methods bring with them diachronic norms that govern how the model should adjust its beliefs and its credences. The details of these norms can be debated, but Jeffrey conditionalization as in (\ref{jc}) is a very plausible candidate. 

Having thus defended both synchronic as well as diachronic norms for belief in language models, it is now time to address the third and final of our initial questions: do present day language models live up to the norms of rationality that apply to them?

\section{Living up to the normative requirements}

The question what norms apply to language models is one that can be discussed rather generally and abstractly. One can argue, as we did above, that particular classes of models are under particular norms of rationality, while others are not. But the question whether a particular model lives up to the norms that apply to it is one that is specific to each model. It depends on how well the model is doing what it is supposed to do, and that depends on the details of the model, and cannot be determined in general. Nonetheless, there is a general question in the neighborhood, which is one about the effectiveness of rational requirements on models in general. Our human minds are under rational norms and somehow these norms are generally effective. Although we are often irrational, it is not the case that rational norms have no effect on us at all. They often push us in the right direction. How rational and irrational human beings are is a widely studied question \citep{tversky-kan1974}. Our concern here is not human beings, but language models. And to determine how rational they are faces a special additional difficulty: Language models face a conflict of norms that does not seem to carry over to human rationality. 

Language models can be under competing normative requirements, and because of this it is hard to see how effective the norms of rationality which apply to them are. For a similar reason, it is hard to see how rational they are when we try to assess the rationality of their beliefs. To illustrate the issue, consider a language model fine-tuned for truthfulness but pretrained on incoherent text. Such a model is under rational coherence norms since it aims at the truth. In particular, it is under the norm to have logically consistent beliefs. To see if it lives up to this coherence norm we can empirically determine if it has consistent beliefs. We can query the model with questions and determine if it gives us logically consistent answers. And thus, it would seem easy to test whether it lives up to the coherence norms that apply to it. 

But unfortunately, it is not that easy. Suppose that we find that the model does not live up to this requirement and has inconsistent beliefs. Is the reason for this that it fails to live up to a normative requirement of logical coherence, which applies to it, and thus it is irrational? That is not so clear, since the model is not only fine-tuned for truthfulness, but also pretrained on incoherent training data, i.e. on text that is not logically consistent. And as a machine learning model, it is also under the requirement to model the training data. However, these two requirements, logical coherence and learning the training data, are in conflict in this case. Being pretrained on incoherent text might support an incoherent set of beliefs as the proper response. But being fine-tuned for truthfulness requires coherent beliefs. If we encounter an inconsistent set of beliefs in a language model, is that because it fails to live up to the rational requirement of coherence or is it because it does live up to the requirement of learning the training data? For language models like the ones we have today, it is hard to separate the two normative forces and settle which one should have stronger pull.  Thus, it is hard to empirically test whether rational norms are effective. 

To solve this problem, and to properly test how effective rational norms are in language models, we need to control one variable to test for the other. In particular, we need to control the incoherence of the training data in pretraining to test for rationality in the fine-tuned models. If we knew, say, that the model was only pretrained on logically consistent training data, then any inconsistency in its beliefs is a result of a rational failure. And if we knew that the training data was only slightly incoherent, then the degree of incoherence in the data might correspond to a degree of irrationality in the model. But, of course, such experiments are not possible with present language models, since they are trained on a large corpus of text that is not controlled for coherence and in fact is known to be vastly incoherent. And what is worse, it is known to be hard to test any large body of natural language text for logical coherence. Just to test for Boolean satisfiability is an NP-complete problem \citep{cook-complexity}. It is likely that pretraining work involving synthetic corpora will be required to determine whether language models produce inconsistencies as a result of adherence to their training data or a failure of rationality.

We take up this question in related empirical work \citep{hase2024fundamentalproblemsmodelediting}. The goal of the experiments there is two-fold. We test (1) whether language models live up to synchronic coherence norms, with respect to a synthetically constructed pretraining corpus, and (2) whether they live up to derivative diachronic norms (see Sec. 4), including the fulfillment of synchronic norms before and after editing a language model to adopt a new fact about the world. 

To test for (1), we pretrain a language model on 1 billion tokens of synthetic data constructed algorithmically according to the rules of a formal language. The corpus consists of sentences about a hypothetical world. There are a limited number of true, atomic propositions about this world. 
We construct a partially incoherent dataset by including true sentences as well as false sentences (sentences that contradict the true sentences) at known frequencies.
Crucially, this enables us to obtain a ``ground truth'' credence for each atomic sentence in the data. In fact, this ``ground truth'' credence is given to us by a Bayesian model, also fit to the incoherent corpus. We will refer to the probabilities learned by this Bayesian model as gold standard credences. Thus, to measure synchronic coherence, we compare the credences of our pretrained language model against these gold standard credences. When we pretrain a language model on our corpus, we find that the resulting language model is quite successful at generating the most common sentences, which are also the ``true'' sentences about our hypothetical world. But the model is quite unsuccessful at giving calibrated probabilities for these sentences. This means that the model's credences are directionally correct, i.e.~they track the truth, but generally imprecise, i.e.~they diverge from the ideal Bayesian gold standard. 

To test for (2), we take this pretrained model and fine-tune it to adopt a new fact about the world, represented by a sentence that contradicts what was seen in the pretraining data. Does this new model maintain a coherent picture of the world? We find that, no, these updates to the model immediately produce some incoherent beliefs. We can say so by assessing the updated language model against gold-standard posterior credences given by our Bayesian model, also fed the same new sentence. The Bayesian model performs exactly the kind of Jeffrey conditionalization given in Eq. \ref{jc}, with the new sentence serving the role of evidence $B$. To give an example of the language model's incoherence, when the model successfully learns the new sentence \emph{P}, it does not then assign the appropriate credence to the sentence ``not \emph{P}''. It also does not assign the right credence to a sentence that is implied by \emph{P} (along the lines of ``\emph{X} is a bachelor'' implying ``\emph{X} is unmarried''). More details are available in our paper \citep{hase2024fundamentalproblemsmodelediting}. 

In sum, this suggests that a pretrained language model's credences only partially track the degree of incoherence in the training data. A model can learn what atomic propositions are true or not, but its credences obtain high Brier scores (i.e., high error). Moreover, coherence norms may sometimes have no effect. After fine-tuning on new sentences, these norms fall quite short of being effective. 

These experiments are conducted in a controlled setting with a synthetic corpus in order to tease apart the role of pretraining data coherence and rational norms. To add context to this result, we note that larger-scale models (LLMs) trained on natural language text often appear to demonstrate a robust understanding of the world. We point to \citep{wilie2024belief} for a study of how such LLMs perform logical inference and belief revision in a more naturalistic setting. We note that LLM abilities show limitations under these quantitative evaluations but have been growing more capable over time.

\section{Conclusion}

To determine whether language models are rational, irrational, or arational, is a crucial part in understanding them, both internally and why they produce the outputs they produce. In this paper we argued that this question does not have a uniform answer for all language models. In particular, we argued that pretrained models generally are arational, since their internal states are solely under the norms of machine learning models to learn the training data. However, fine-tuned models can be under rational norms, as long as they are fine-tuned for truthfulness, which is the case for models fine-tuned with RLHF, or on special datasets selected for truthfulness. Such models are under both logical and probabilistic coherence norms. With regard to the latter, we proposed that the notion of a credence makes sense for language models, and that we can define a coherent credence function for a model from next token probabilities, exploiting the minimal assent connection MAC. As for how well language models live up to coherence norms, we leave this question to future work studying model coherence with the proper controls for the consistency of the training data.

Overall, this work points in the direction that considerations of rationality do apply to language models, although the difference of being trained on incoherent data in pretraining, and not a coherent world directly, makes an important contribution to understanding how rational norms have a different pull in language models than in humans. And even if it is not clear how well language models live up to coherence norms in practice, we can at least employ our tools of predicting and explaining behavior in terms of representational states for understanding both humans and models.

\addcontentsline{toc}{section}{Bibliography}

\bibliographystyle{acl_natbib}

\bibliography{rational}

@article{joyce:prob,
	author = {James Joyce},
	journal = {Philosophy of Science},
	number = {4},
	pages = {575 - 603},
	title = {A non-pragmatic vindictation of probabilism},
	volume = {65},
	year = {1998}}

@article{bender-koller:climbing,
	author = {Emily Bender and Alexander Koller},
	journal = {Proceedings of the 58th Annual Meeting of the Association for Computational Linguistics},
	pages = {5185--5198},
	title = {Climbing towards {NLU}: On Meaning, Form, and Understanding in the Age of Data},
	year = {2020}}

@incollection{williams:deciding,
	author = {Bernard Williams},
	booktitle = {Problems of the Self},
	pages = {136 - 51},
	publisher = {Cambridge University Press},
	title = {Deciding to believe},
	year = {1973}}

@book{worsnip:fitting,
	author = {Alex Worsnip},
	publisher = {Oxford University Press},
	title = {Fitting Things Together: Coherence and the Demands of Structural Rationality},
	year = {2021}}

@book{teitelbaum:bayes,
	author = {Michael Teitelbaum},
	publisher = {Oxford University Press},
	title = {Fundamentals of Bayesian Epistemology},
	volume = {1 and 2},
	year = {2023}}

@article{Piantadosi-hill,
	author = {Steven T. Piantadosi and Felix Hill},
     archiveprefix = {arXiv},
    eprint = {2208.02957v2},
	journal = {arXiv preprint arXiv:2208.02957v2},
	title = {Meaning without reference in large language models},
	year = {2022}}

@incollection{velleman:aim,
	author = {J. David Velleman},
	booktitle = {The Possibility of Practical Reason},
	pages = {244--81},
	publisher = {Oxford University Press},
	title = {On the aim of belief},
	year = {2000}}

@article{russell:rational,
	author = {Stuart Russell},
	journal = {Artificial Intelligence},
	pages = {57 - 77},
	title = {Rationality and intelligence},
	volume = {94},
	year = {1997}}

@book{kiesewetter:norm,
	author = {Benjamin Kiesewetter},
	publisher = {Oxford University Press},
	title = {The Normativity of Rationality},
	year = {2017}}

@article{railton:truth,
	author = {Peter Railton},
	journal = {Philosophical Issues},
	number = {71-93},
	title = {Truth, reason, and the regulation of belief},
	volume = {5},
	year = {1994}}

@article{kolodny:rational,
	author = {Niko Kolodny},
	journal = {Mind},
	number = {455},
	pages = {509-563},
	title = {Why be rational?},
	volume = {114},
	year = {2005}}

@inproceedings{de2021editing,
	title        = {Editing Factual Knowledge in Language Models},
	author       = {De Cao, Nicola  and Aziz, Wilker  and Titov, Ivan},
	year         = 2021,
	month        = nov,
	booktitle    = {EMNLP},
	publisher    = {Association for Computational Linguistics},
	pages        = {6491--6506},
	url          = {https://aclanthology.org/2021.emnlp-main.522}
}

@article{dai2021knowledge,
	title        = {Knowledge neurons in pretrained transformers},
	author       = {Dai, Damai and Dong, Li and Hao, Yaru and Sui, Zhifang and Wei, Furu},
	year         = 2021,
	journal      = {arXiv preprint arXiv:2104.08696},
	url          = {https://arxiv.org/pdf/2104.08696.pdf}
}

@article{zhu2020modifying,
	title        = {Modifying Memories in Transformer Models},
	author       = {Zhu, Chen and Rawat, Ankit Singh and Zaheer, Manzil and Bhojanapalli, Srinadh and Li, Daliang and Yu, Felix and Kumar, Sanjiv},
	year         = 2020,
	journal      = {arXiv preprint arXiv:2012.00363},
	url          = {https://arxiv.org/pdf/2012.00363.pdf}
}

@article{hase2021language,
	title        = {Do language models have beliefs? methods for detecting, updating, and visualizing model beliefs},
	author       = {Hase, Peter and Diab, Mona and Celikyilmaz, Asli and Li, Xian and Kozareva, Zornitsa and Stoyanov, Veselin and Bansal, Mohit and Iyer, Srinivasan},
	year         = 2021,
	journal      = {arXiv preprint arXiv:2111.13654},
	url          = {https://arxiv.org/pdf/2111.13654.pdf}
}

@article{zhou-relying,
	title        = {Relying on the Unreliable: The Impact of Language Models' Reluctance to Express Uncertainty},
	author       = {Zhou, Kaitlyn and Hwang, Jena and Ren, Xiang and Sap, Maarten},
	year         = 2024,
	journal      = {arXiv preprint arXiv:2401.06730},
	url          = {https://arxiv.org/pdf/2401.06730v1}
	}

@article{huang2023reasoning,
      title={Towards Reasoning in Large Language Models: A Survey}, 
      author={Jie Huang and Kevin Chen-Chuan Chang},
      year={2023},
    journal = {arXiv preprint arXiv:2212.10403},
      eprint={2212.10403},
      archivePrefix={arXiv},
      primaryClass={cs.CL},
      url={https://arxiv.org/abs/2212.10403}
}

@InCollection{sep-knowledge-analysis,
	author       =	{Ichikawa, Jonathan Jenkins and Steup, Matthias},
	title        =	{{The Analysis of Knowledge}},
	booktitle    =	{The {Stanford} Encyclopedia of Philosophy},
	editor       =	{Edward N. Zalta},
	howpublished =	{\url{https://plato.stanford.edu/archives/sum2018/entries/knowledge-analysis/}},
	year         =	{2018},
	edition      =	{{S}ummer 2018},
	publisher    =	{Metaphysics Research Lab, Stanford University}
}

@article{tversky-kan1974,
	author = {Amos Tversky and Daniel Kahneman},
	journal = {Science},
	number = {4157},
	pages = {1124-1131},
	title = {Judgment under Uncertainty: Heuristics and Biases},
	volume = {185},
	year = {1974}}

@article{cook-complexity,
	author = {Stephen Cook},
	journal = {STOC '71: Proceedings of the third annual ACM symposium on Theory of computing},
	pages = {151--158},
	title = {The complexity of theorem-proving procedures},
	year = {1971}}

@misc{mollo2025vectorgroundingproblem,
      title={The Vector Grounding Problem}, 
      author={Dimitri Coelho Mollo and Raphaël Millière},
      year={2025},
      eprint={2304.01481},
      archivePrefix={arXiv},
      primaryClass={cs.CL},
      url={https://arxiv.org/abs/2304.01481}, 
}

@article{hase2024fundamentalproblemsmodelediting,
      title={Fundamental Problems With Model Editing: How Should Rational Belief Revision Work in {LLM}s?}, 
      author={Peter Hase and Thomas Hofweber and Xiang Zhou and Elias Stengel-Eskin and Mohit Bansal},
      year={2024},
    journal = {Transactions of Machine Learning Research},
      url={https://openreview.net/pdf?id=LRf19n5Ly3}, 
}

@article{wilie2024belief,
  title={Belief revision: The adaptability of large language models reasoning},
  author={Wilie, Bryan and Cahyawijaya, Samuel and Ishii, Etsuko and He, Junxian and Fung, Pascale},
  journal={arXiv preprint arXiv:2406.19764},
  year={2024},
  url={https://arxiv.org/pdf/2406.19764?}
}

@book{fodor1987psychosemantics,
  title={Psychosemantics: The Problem of Meaning in the Philosophy of Mind},
  author={Fodor, Jerry A.},
  year={1987},
  publisher={MIT Press},
  address={Cambridge, MA}
}

@book{brentano1995psychology,
  title={Psychology from an Empirical Standpoint},
  author={Brentano, Franz},
  year={1995},
  publisher={Routledge},
  address={London},
  editor={McAlister, Linda L.},
  translator={Rancurello, Antos C. and Terrell, D. B. and McAlister, Linda L.},
  note={Originally published 1874}
}

@book{skinner1957verbal,
  title={Verbal Behavior},
  author={Skinner, B. F.},
  year={1957},
  publisher={Appleton-Century-Crofts},
  address={New York}
}

@book{ryle1949concept,
  title={The Concept of Mind},
  author={Ryle, Gilbert},
  year={1949},
  publisher={Hutchinson},
  address={London}
}

@book{grice:studies,
	author = {Paul Grice},
	publisher = {Harvard University Pressr},
	title = {Studies in the Way of Words},
	year = {1989}
}

@book{austin1962how,
  title={How to Do Things with Words},
  author={Austin, J. L.},
  year={1962},
  publisher={Harvard University Press},
  address={Cambridge, MA},
  note={The William James Lectures delivered at Harvard University in 1955}
}

@inproceedings{lorini:design,
author = {Lorini, Emiliano},
title = {Designing Artificial Reasoners for Communication},
year = {2024},
isbn = {9798400704864},
publisher = {International Foundation for Autonomous Agents and Multiagent Systems},
address = {Richland, SC},
booktitle = {Proceedings of the 23rd International Conference on Autonomous Agents and Multiagent Systems},
pages = {2690–2695},
numpages = {6},
location = {Auckland, New Zealand},
series = {AAMAS '24}
}

@article{alchourron1985logic,
  title={On the logic of theory change: Partial meet contraction and revision functions},
  author={Alchourr{\'o}n, Carlos E. and G{\"a}rdenfors, Peter and Makinson, David},
  journal={The Journal of Symbolic Logic},
  volume={50},
  number={2},
  pages={510--530},
  year={1985},
  publisher={Cambridge University Press}
}

\end{document}